\documentclass[letterpaper, 10 pt, conference]{ieeeconf}  % Comment out if you need a4paper

\IEEEoverridecommandlockouts
\usepackage[T1]{fontenc}
\usepackage{graphicx}
\usepackage{subcaption}
\usepackage{svg}
\usepackage{amsmath,amssymb}
\usepackage{siunitx}
\usepackage{verbatim}

\title{\LARGE \bf
Cognitive and Behavioural Effects of Haptic Feedback \\ for ROV Teleoperation in Hazardous Environments
}

\author{Joshua Brown$^{1}$ \and Ildar Farkhatdinov$^{2}$ \and Michael Jenkin$^{3}$% <-this % stops a space
\thanks{$^{1}$Department of Surgery and Cancer, Imperial College London, London, UK.}
\thanks{$^{2}$School of Biomedical Engineering and Imaging Sciences, King's College London, London, UK.}
\thanks{$^{3}$School of Electrical Engineering and Computer Science, York University, Toronto, ON, Canada.}}%

\begin{document}

\maketitle
\thispagestyle{empty}
\pagestyle{empty}

%%%%%%%%%%%%%%%%%%%%%%%%%%%%%%%%%%%%%%%%%%%%%%%%%%%%%%%%%%%%%%%%%%%%%%%%%%%%%%%%
\begin{abstract}
Teleoperating a Remotely Operated Vehicle (ROV) in complex underwater environments presents unique challenges. Dynamic, three dimensional obstacles and unseen water currents pose real risks to expensive robotic hardware in environments in which recovery can be next to impossible. This paper explores the cognitive effects of haptic feedback to improve vehicle safety in such environments. An in-water study is analysed with a ROV exposed to controlled cross-currents, with environmental information presented via a soft and vibrating haptic touchpad.

Results demonstrate that while encountering an unseen cross-current significantly increases an operator's physical and temporal demand, the introduction of haptic feedback mitigates these factors and improves self-assessed performance. Interestingly, the haptic condition also produced the highest levels of mental demand and operator frustration. Despite this increased mental workload, operators drove the ROV significantly faster when receiving haptic cues compared to the current-only condition, indicating a higher level of user confidence. Ultimately, these findings suggest that while haptic feedback may initially require more cognitive processing to interpret unseen hazards, it successfully enhances operator awareness and promotes more confident navigation in hazardous environments.
\end{abstract}

%%%%%%%%%%%%%%%%%%%%%%%%%%%%%%%%%%%%%%%%%%%%%%%%%%%%%%%%%%%%%%%%%%%%%%%%%%%%%%%%
\section{Introduction}

% Problem statement
Teleoperating a robot underwater carries significant and unique challenges. Unlike ground robots, ROVs can encounter dynamic hazards from any direction in 3D space, which may or may not be visible to onboard sensors or cameras. Unlike drones, visibility underwater is notoriously poor, with inconsistent lighting, high turbidity, and visual distortions among the many problems associated with operating in this environment~\cite{smiles2025robotic,Smiles2026}. Haptic feedback has been proposed as a means of delivering denser spatial information to the operators of remote robots, however the effects of this additional information on cognitive load and confidence remain broadly unexplored.

Haptic communication of environmental and robot state information is an increasingly well-investigated area. It is known that force feedback can be used to provide directional cues to warn of hazards~\cite{Tang2009HapticReality,farkhatdinov2008control} or stimulate flight instability~\cite{Rognon2019SoftDrone}. Haptic feedback has also been used to enhance communication between humans and robots in shared control paradigms, particularly in underwater environments \cite{Sakagami2022}. Other examples of haptic feedback being used to enhance underwater teleoperation include use of force feedback to restrict access to hazardous `keep-out' zones~\cite{Ryden2013AdvancedRendering} and mitigate the effects of satellite delay in long-distance teleoperation~\cite{DiLillo2021UnderwaterProject}.
For a full and recent review of the applications of haptic feedback to robot teleoperation, see \cite{Pacchierotti2024Cutaneous/TactilePerspectives}.

\begin{figure}
\centering
\includegraphics[width=\columnwidth]{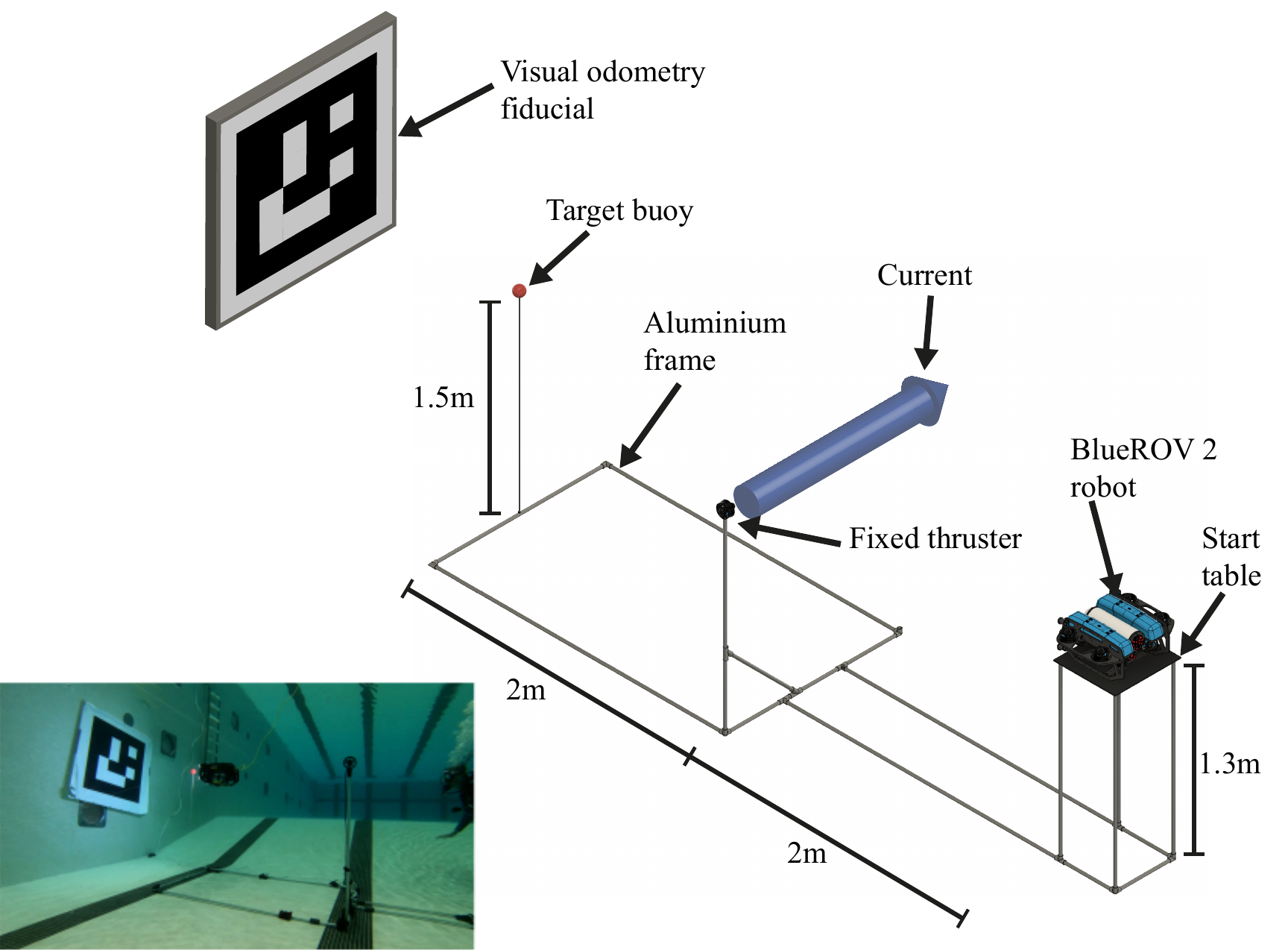}
\caption{The experimental setup showing the ROV and target buoy, separated by a controllable cross-current.}
\label{fig:setup}
\end{figure}

In a recent in-water study, we showed that multimodal haptic feedback comprising vibration and softness-change can improve the quantitative positioning accuracy, and path efficiency of operators of ROVs in the presence of invisible cross-currents~\cite{Brown2025ROVHaptics}. This paper extends these previously published quantitative performance indicators with a new exploration of the effect of haptic feedback on operator confidence and the specific components of cognitive load. We present a controlled, in-water study investigating the behavioural and cognitive effects of haptic feedback during hazardous teleoperation scenarios. 
%The main contribution is an experimental assessment of the cognitive and behavioural effects of haptic feedback on ROV operators in hazardous conditions.

\section{Experimental Study}

An experiment was devised in which novice participants teleoperated a BlueRobotics ROV in a straight line path, toward a visible target, using visual information provided by a camera mounted on the ROV. A controllable cross-current was introduced in the middle of the planned path to provide a hazard to vehicle control. In addition to Key Performance Indicators (KPIs), users' cognitive load was assessed via the well-established NASA TLX questionnaire~\cite{Hart1988DevelopmentResearch}, whilst the robot's average speed was calculated as a recognised proxy for confidence ~\cite{Schmidt-Daffy2013FearMotivation,Taylor2006StressInactivity}.

\subsection{Haptic feedback and interaction}
\begin{figure}
\centering
\includegraphics[width=\columnwidth]{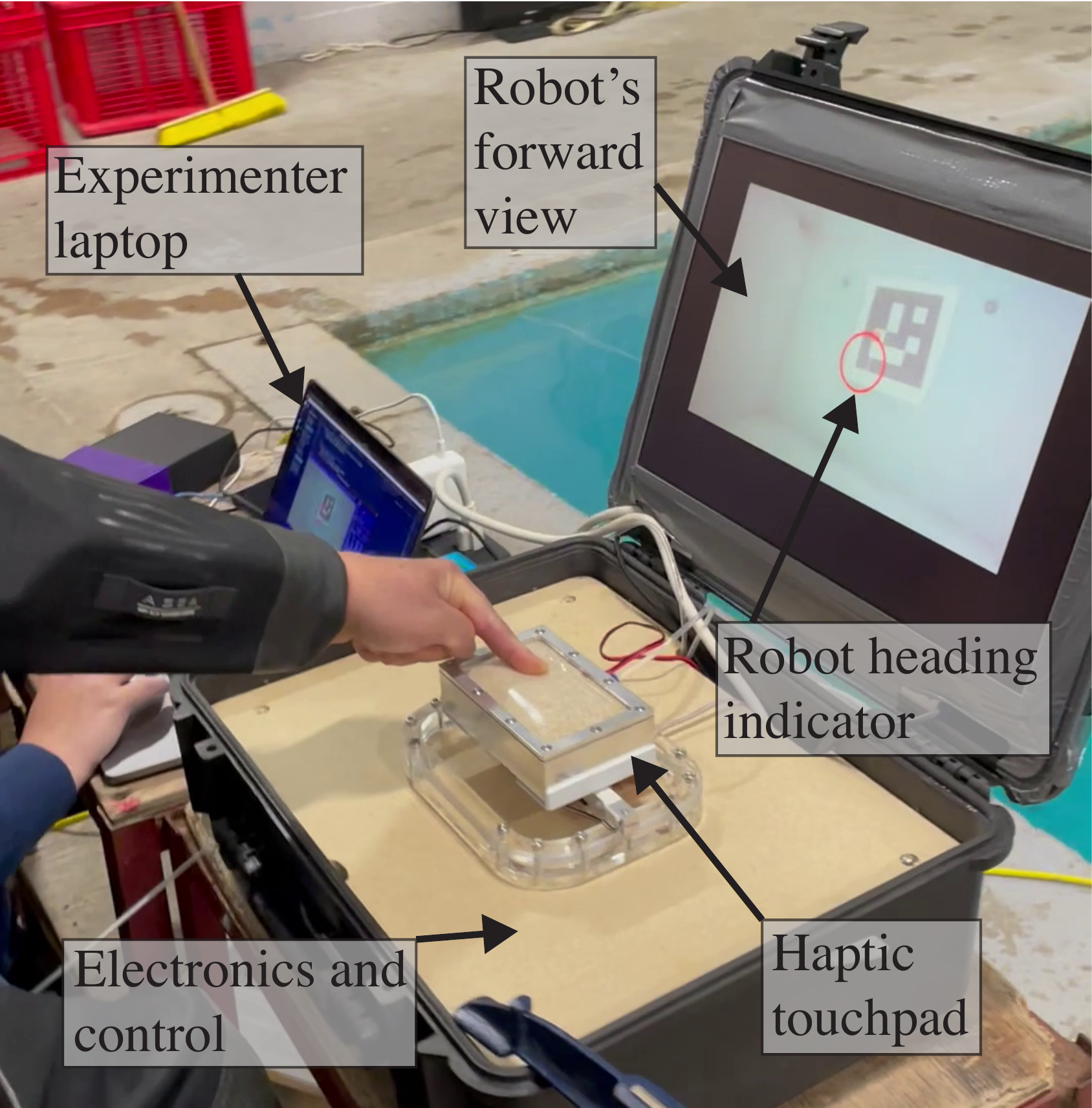}
\caption{The operator's control station with a view from the robot's front camera and the soft touchpad.}
\label{fig:interface}
\end{figure}

Haptic feedback was provided through a planar haptic touchpad in two dimensions. Softness-change, effected via particle jamming~\cite{Follmer2012}, was combined with vibration as described in~\cite{Brown2020_Soft_Haptic_Interface_Based_On_Vibration} to offer two distinct haptic cues. Softness was used to indicate distance to the target buoy, whilst an asymmetric vibration (produced using an approach similar to~\cite{Tappeiner2009}) indicated the presence and direction of the current. Both of these cues were delivered using a haptic touchpad, described in detail in~\cite{Brown2021Shape-ChangingVibration}. The operator station is shown in Figure~\ref{fig:interface}.

Human-operator's input was taken by a force-sensing base built into the touchpad. Load cells on the left and right recorded the users' left-right finger position for steering, and magnitude of force for speed (press harder, go faster).

\subsection{Participants}
Eight participants (five male) between 20 and 70 years of age were recruited to take part in the study, which received ethical approval from York University (certificate e2022-266). Five individuals reported regularly playing video games, and one reported regularly teleoperating RC vehicles. All participants had normal or corrected-to-normal vision and reported no impairments to their sense of touch or fine motor control.

\subsection{Setup}

Participants were asked to teleoperate an ROV (BlueRobotics BlueROV2-Heavy) ahead 4~m to a visible red buoy. A fixed thruster was positioned such that a left-right cross-current could be created in the middle of the robot's path. All experiments took place in controlled, indoor pools. The robot and buoy were fixed at a constant and identical depth by automated depth control provided by the ROV flight controller, and a string anchoring the buoy to the bottom of the pool. A large fiducial marker was positioned at the end of the robot's path well beyond the buoy to enable visual odometry. A custom-built aluminium fame was used to maintain the positions of underwater apparatus during the experiment. The in-water setup is shown in Fig. \ref{fig:setup}.

\subsection{Protocol}

Participants undertook the task in three conditions. In the control condition, no cross-current was present (still water) and no haptic feedback was provided. In the cross-current condition, the current was switched on with no haptic feedback provided. In the haptics condition, the cross-current was present and haptic feedback was used to indicate its presence when encountered. Each condition was delivered five times, with blocks of five presented in a random sequence. A second sequence of three blocks of five trials was then presented to assess whether participants performed differently after being given an opportunity to familiarise themselves with the robot.

\section{Results}

The results were tested for normality via a Jarque-Bera test. As these were found to be normally distributed, conditions were compared using post-hoc two-tailed paired t-tests. All statistical tests were performed using standard tools available in the Matlab R2025b Statistics and Machine Learning Toolbox. Outliers were defined as responses more than 1.5 times the  interquartile range below the 25th percentile or above the 75th percentile and were removed from figures for clarity.

\subsection{Cognitive Load}
\begin{figure}[t]
\centering
\includegraphics[width=\columnwidth]{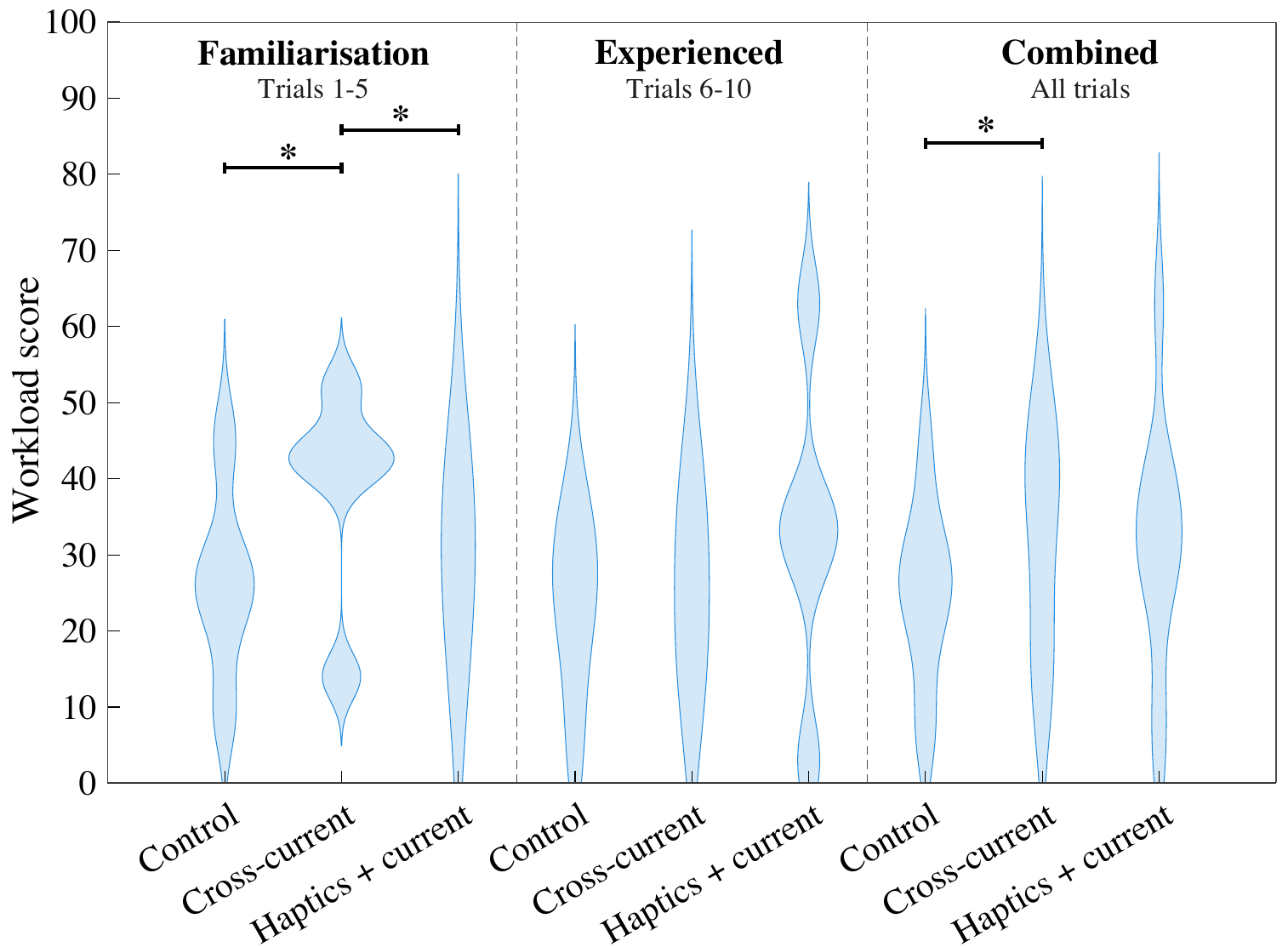}
\caption{NASA TLX score (cognitive load) reported by the robot operator in each condition and phase.}
\label{fig:load}
\end{figure}

Overall, cognitive load as measured using the NASA TLX score was found to be significantly increased by the introduction of the cross-current (32.07 overall; familiarisation: 38.73; experienced: 25.40) compared to the control (24.83 overall; familiarisation: 26.40; experienced: 23.27) ($p$=0.01487), however was only slightly reduced by the introduction of haptic feedback (31.31 overall; familiarisation: 29.27; experienced: 33.00). A graphical comparison is shown in Figure \ref{fig:load}.

\begin{figure}
\centering
\includegraphics[width=\columnwidth]{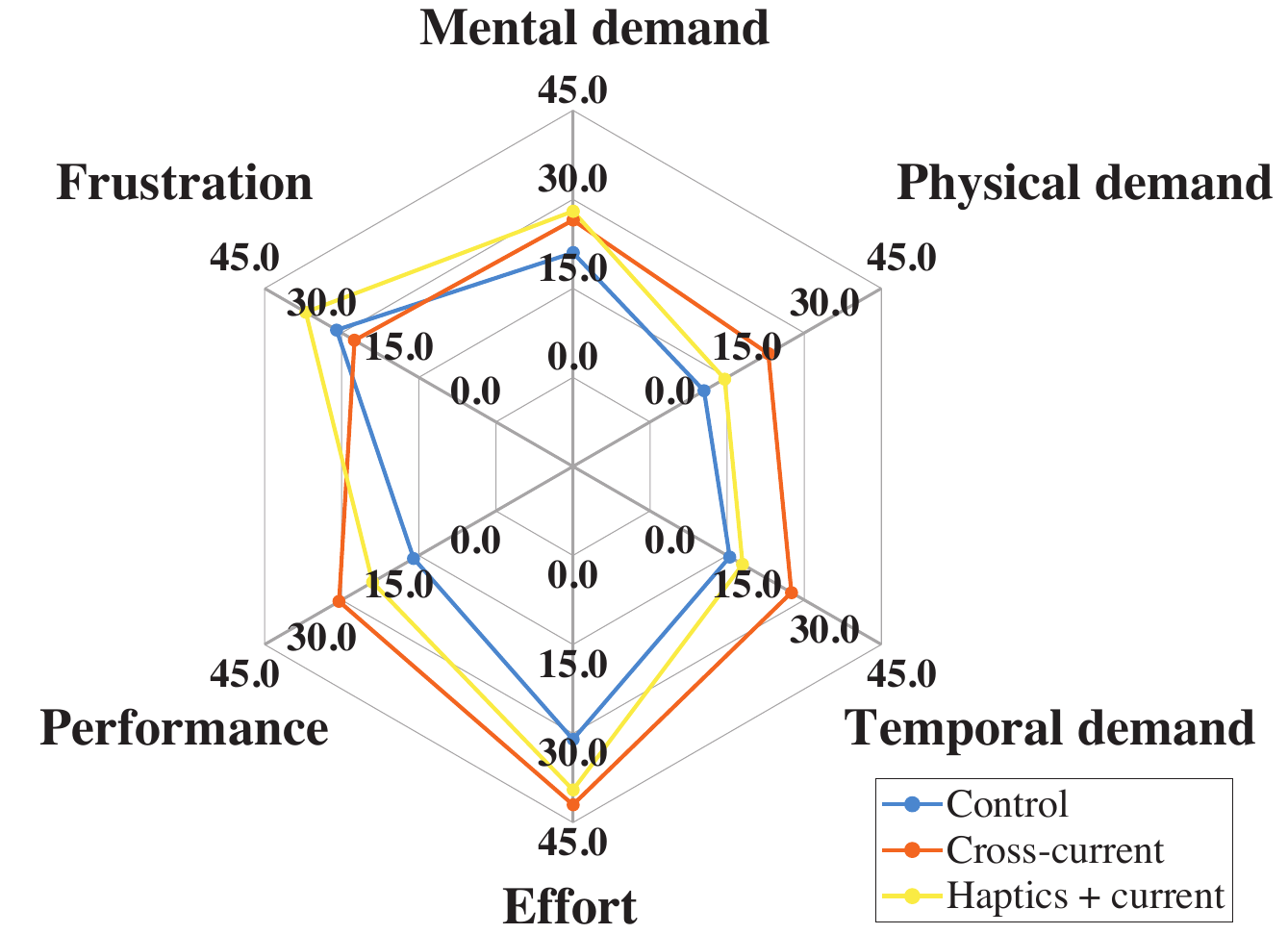}
\caption{Comparison of cognitive load components (raw scores before participant weighting, with the performance scale reversed to enable comparison between components).}
\label{fig:load-breakdown}
\end{figure}

As well as an overall score, NASA-TLX also reports cognitive load across six individual components (mental demands, physical demand, temporal demand, effort, performance and frustration). Results show that the cross-current increased the task's physical and temporal demand, as well as perceived effort. Self-assessed performance was also worsened in the presence of the cross-current. These components were all improved with the introduction of haptic feedback, but not to the level of the control condition. Mental demand was highest in the haptic condition, closely followed by the cross-current condition. Surprisingly, the haptic condition was also judged to be the most frustrating, followed by the control, with the cross-current least frustrating. Full results are shown in Table~\ref{tab:rov-tlx-table} and compared graphically in Figure~\ref{fig:load-breakdown}.


\begin{table}[b]
\centering
\begin{tabular}{lrrr}
\textbf{}          & \multicolumn{3}{c}{\textbf{Mean}}                                                                                        \\ \cline{2-4} 
\textbf{Component} & \multicolumn{1}{l}{\textbf{Control}} & \multicolumn{1}{l}{\textbf{Cross-current}} & \multicolumn{1}{l}{\textbf{Haptics}} \\ \hline
Mental demand      & 21±14.3                       & 26.5±19.6                       & 28±24.5                                   \\ %\hline
Physical demand    & 10.5±10.1                     & 23±19.7                         & 14.5±13.6                                 \\ %\hline
Temporal demand    & 15.5±10.7                     & 27.5±23.7                       & 18±15.3                                   \\ %\hline
Effort             & 31±16.5                       & 42±17.7                         & 39.5±20.6                                 \\ %\hline
Performance        & 16±13.1                       & 30.5±22.8                       & 24±19.4                                   \\ %\hline
Frustration        & 31±24.7                       & 27.5±20.2                       & 37±29.6                                  
\end{tabular}
\caption{Mean and standard deviation of NASA-TLX components in each condition. Note that the performance scale is inverted, such that higher numbers indicate higher perceived performance rather than higher workload.}
\label{tab:rov-tlx-table}
\end{table}

\subsection{Operator confidence}

Past research has demonstrated that higher self-confidence of vehicle operators can manifest as a tendency to use higher speeds~\cite{Schmidt-Daffy2013FearMotivation,Taylor2006StressInactivity}. For this reason, average speed was calculated based on the robot's ground speed log and used as an indirect indicator of confidence. Participants were found to teleoperate the ROV significantly faster with haptic feedback (0.388~m$\cdot$s$^{-1}$) than in the cross-current condition (0.291~m$\cdot$s$^{-1}$) ($p$=0.01619) or control (0.362~m$\cdot$s$^{-1}$), indicating higher confidence.
%Ildar: not clear to me what is p value related to? if it is significant, why p is higer?

\begin{figure}
\centering
\includegraphics[width=\columnwidth]{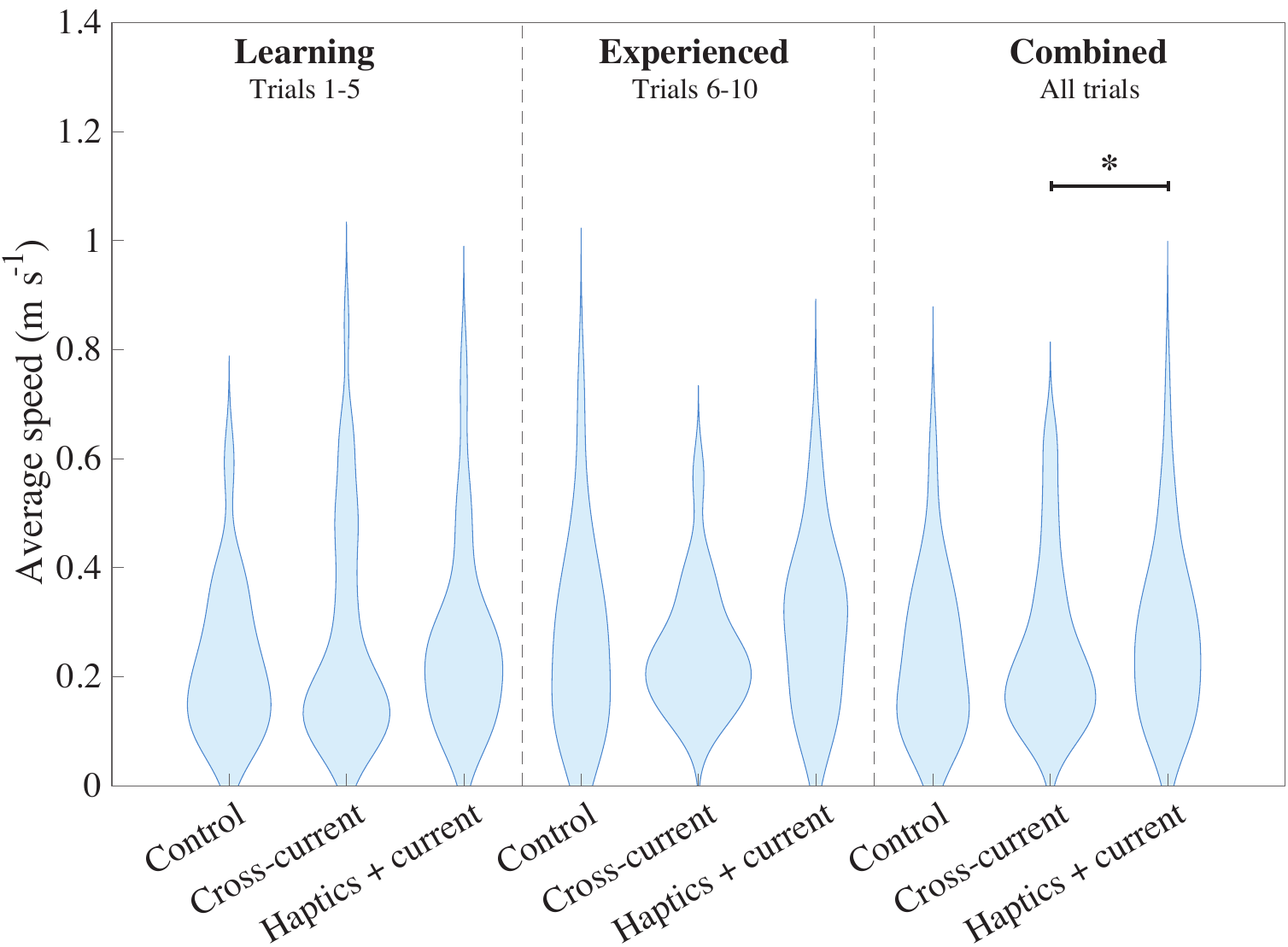}
\caption{Average speed of the robot in each condition and phase with outliers removed.}
\label{fig:speed}
\end{figure}

\section{Discussion and Conclusion}

Whilst the previous analysis of this experiment in~\cite{Brown2025ROVHaptics} focused strictly on quantitative task performance (aggregate cognitive load, position accuracy and path efficiency), demonstrating that haptic feedback improved novice positioning accuracy by 28.3\% and reduced path deviation for experienced operators by 29.5\%, it shed little light on the complex psychological impacts of these interventions. This paper begins to address this gap and takes a materially different focus, investigating the nuanced behavioural responses, confidence levels, and multi-dimensional cognitive load experienced by the human operator.

The overall reduction on cognitive load with haptic feedback observed here was expected and the result is consistent with a substantial body of literature such as~\cite{Rahal2020CaringTelemanipulation}. However, it was not expected that the provision of haptic feedback would make the task more frustrating or mentally demanding. This may be due to the enhanced provision of environmental information, making the operator more aware of the unseen hazard and loss of navigational control. This improved awareness would also explain the improved task performance with haptic feedback noted in~\cite{Brown2025ROVHaptics}.

Correlating with the apparent increase in confidence indicates that whilst participants reported that the task was more frustrating, participants did have more confidence with conditions providing haptic feedback. This would align with the improved awareness of the hazard -- it was frustrating, but once understood could be addressed -- and improved performance observed and discussed previously, especially in the familiarisation phase when the presence of the unseen hazard would have been most surprising. Moreover between the learning and experienced phases, average speeds increased slightly across all participants, and the highest speeds increased substantially in the control and haptics conditions, indicating increased comfort. Whilst speeds decreased slightly on the cross-current condition in the experienced phase indicating that participants became increasingly wary of the cross-current when they encountered it without the explanation provided by the haptic cue.

Future work should investigate this interplay further. A simulation study would provide a more scalable environment for further research, with a broader range of visual and haptic conditions used to tease out changes in the operator's behaviour and psychological state. Real-time monitoring of cognitive load, not possible with the post-hoc NASA-TLX questionnaire, would also be beneficial to study correlations with the haptic feedback. Enhancement of the haptic device would also be possible with more recent advances in soft and multi-modal haptic technologies, with pump-free particle jamming offering better integration with other modalities and faster response times than were possible when this study was conducted~\cite{Brown2024DesignActuators,Brown2026AModulation}.

Finally, whilst this work engaged novice ROV operators in a scientific environment, it is important to contextualise the work in real-world ROV operations. In these cases, missions would usually be conducted by experienced controllers. This stronger baseline experience may reduce the initial cognitive load spikes and allow the benefits of the haptic feedback to be gained faster. Further research with expert operators should be conducted to investigate this effect. Furthermore, while this study successfully simulated a hazard using a fixed thruster to generate a predictable cross-current in a controlled, indoor pool, real-world underwater environments are considerably more volatile. In true field applications, visibility is notoriously poor due to high turbidity, visual distortions, and inconsistent lighting. Under such conditions, visual information would likely be obscured or completely fail. As the primary visual channel degrades, the value of the multimodal haptic feedback tested here would theoretically increase, providing a helpful sensory redundancy. Future work must evaluate this haptic framework against dynamic, unpredictable 3D hazards to fully understand its potential to safeguard expensive robotic hardware in truly extreme environments.

%\section{Conclusion}

When teleoperating a remote vehicle in hazardous environments, reporting information to the operator is vital. This work began to explore the psychological interplay between awareness of hazards and ability to respond to them, demonstrating that providing this information through the haptic channel has clear advantages not just to performance, but also to awareness and opportunity to adapt.

%%%%%%%%%%%%%%%%%%%%%%%%%%%%%%%%%%%%%%%%%%%%%%%%%%%%%%%%%%%%%%%%%%%%%%%%%%%%%%%%
% Appendices should appear before the acknowledgment in the IEEE template
%\appendix

%%%%%%%%%%%%%%%%%%%%%%%%%%%%%%%%%%%%%%%%%%%%%%%%%%%%%%%%%%%%%%%%%%%%%%%%%%%%%%%%
\section*{ACKNOWLEDGMENT}

This work was partly funded by UKRI (NE/V010018/1) and Mitacs under the Globalink UK-Canada Doctoral Exchange program.

The authors thank Geoff Simpson of Queen Mary University of London, Chris Giverin of York University, James Zacher of Ronin Commercial Services and Heather Jenkin of York University for technical and operational support.

%\section*{DISCLOSURE OF INTERESTS}
%The authors have no competing interests to declare.

%%%%%%%%%%%%%%%%%%%%%%%%%%%%%%%%%%%%%%%%%%%%%%%%%%%%%%%%%%%%%%%%%%%%%%%%%%%%%%%%
% Bibliography
% If you already have a .bib file named joshua.bib, this will work.
% IEEE typically uses IEEEtran.bst.
\bibliographystyle{IEEEtran}
\bibliography{joshua}

\begin{thebibliography}{10}
\providecommand{\url}[1]{#1}
\csname url@samestyle\endcsname
\providecommand{\newblock}{\relax}
\providecommand{\bibinfo}[2]{#2}
\providecommand{\BIBentrySTDinterwordspacing}{\spaceskip=0pt\relax}
\providecommand{\BIBentryALTinterwordstretchfactor}{4}
\providecommand{\BIBentryALTinterwordspacing}{\spaceskip=\fontdimen2\font plus
\BIBentryALTinterwordstretchfactor\fontdimen3\font minus \fontdimen4\font\relax}
\providecommand{\BIBforeignlanguage}[2]{{%
\expandafter\ifx\csname l@#1\endcsname\relax
\typeout{** WARNING: IEEEtran.bst: No hyphenation pattern has been}%
\typeout{** loaded for the language `#1'. Using the pattern for}%
\typeout{** the default language instead.}%
\else
\language=\csname l@#1\endcsname
\fi
#2}}
\providecommand{\BIBdecl}{\relax}
\BIBdecl

\bibitem{smiles2025robotic}
A.~Smiles, C.~Oh, and I.~Farkhatdinov, ``Robotic perception of underwater plastic bottles for augmented telepresence,'' in \emph{European Robotics Forum}.\hskip 1em plus 0.5em minus 0.4em\relax Springer, 2025, pp. 320--326.

\bibitem{Smiles2026}
A.~L. Smiles, C.~Oh, and I.~Farkhatdinov, ``Descriptor: Underwater stereo vision dataset for transparent object detection \& ranging (uw-transstereo),'' \emph{IEEE Data Descriptions}, vol.~3, pp. 638--645, 2026.

\bibitem{Tang2009HapticReality}
X.~Tang, D.~Zhao, H.~Yamada, and T.~Ni, ``{Haptic interaction in tele-operation control system of construction robot based on virtual reality},'' \emph{2009 IEEE International Conference on Mechatronics and Automation, ICMA 2009}, pp. 78--83, 2009.

\bibitem{farkhatdinov2008control}
I.~Farkhatdinov, J.~H. Ryu, and J.~Poduraev, ``Control strategies and feedback information in mobile robot teleoperation,'' \emph{IFAC Proceedings Volumes}, vol.~41, no.~2, pp. 14\,681--14\,686, 2008.

\bibitem{Rognon2019SoftDrone}
C.~Rognon, M.~Koehler, C.~Duriez, D.~Floreano, and A.~M. Okamura, ``{Soft haptic device to render the sensation of flying like a drone},'' \emph{IEEE Robot. Autom. Lett.}, vol.~4, no.~3, pp. 2524--2531, 7 2019.

\bibitem{Sakagami2022}
N.~Sakagami, M.~Suka, Y.~Kimura, E.~Sato, and T.~Wada, ``{Haptic shared control applied for ROV operation support in flowing water},'' \emph{Artificial Life and Robotics}, vol.~27, no.~4, pp. 867--875, 2022.

\bibitem{Ryden2013AdvancedRendering}
F.~Ryden, A.~Stewart, and H.~J. Chizeck, ``{Advanced telerobotic underwater manipulation using virtual fixtures and haptic rendering},'' in \emph{OCEANS 2013}.\hskip 1em plus 0.5em minus 0.4em\relax San Diego: MTS/IEEE, 2013.

\bibitem{DiLillo2021UnderwaterProject}
P.~Di~Lillo, E.~Simetti, F.~Wanderlingh, G.~Casalino, and G.~Antonelli, ``{Underwater Intervention with Remote Supervision via Satellite Communication: Developed Control Architecture and Experimental Results within the Dexrov Project},'' \emph{EEE Trans. Control Syst. Technol.}, vol.~29, no.~1, pp. 108--123, 2021.

\bibitem{Pacchierotti2024Cutaneous/TactilePerspectives}
C.~Pacchierotti and D.~Prattichizzo, ``{Cutaneous/Tactile Haptic Feedback in Robotic Teleoperation: Motivation, Survey, and Perspectives},'' \emph{IEEE Transactions on Robotics}, vol.~40, pp. 978--998, 2024.

\bibitem{Brown2025ROVHaptics}
J.~Brown, I.~Farkhatdinov, and M.~Jenkin, ``{ROV Teleoperation in the Presence of Cross‐Currents Using Soft Haptics},'' \emph{Journal of Field Robotics}, vol.~42, no.~6, pp. 2580--2593, 9 2025.

\bibitem{Hart1988DevelopmentResearch}
S.~G. Hart and L.~E. Staveland, ``{Development of NASA-TLX (Task Load Index): Results of Empirical and Theoretical Research},'' \emph{Advances in Psychology}, vol.~52, no.~C, pp. 139--183, 1 1988.

\bibitem{Schmidt-Daffy2013FearMotivation}
\BIBentryALTinterwordspacing
M.~Schmidt-Daffy, ``{Fear and anxiety while driving: Differential impact of task demands, speed and motivation},'' \emph{Transportation Research Part F: Traffic Psychology and Behaviour}, vol.~16, pp. 14--28, 2013. [Online]. Available: \url{http://dx.doi.org/10.1016/j.trf.2012.07.002}
\BIBentrySTDinterwordspacing

\bibitem{Taylor2006StressInactivity}
A.~H. Taylor and L.~Dorn, ``{Stress, fatigue, health, and risk of road traffic accidents among professional drivers: The contribution of physical inactivity},'' \emph{Annu. Rev. Public Health}, vol.~27, no.~16, pp. 371--391, 2006.

\bibitem{Follmer2012}
S.~Follmer, D.~Leithinger, A.~Olwal, N.~Cheng, and H.~Ishii, ``{Jamming user interfaces: Programmable particle stiffness and sensing for malleable and shape-changing devices},'' \emph{ACM UIST'12}, pp. 519--528, 2012.

\bibitem{Brown2020_Soft_Haptic_Interface_Based_On_Vibration}
J.~P. Brown and I.~Farkhatdinov, ``{Soft Haptic Interface based on Vibration and Particle Jamming},'' in \emph{IEEE Haptics Symposium, HAPTICS}, vol. 2020-March.\hskip 1em plus 0.5em minus 0.4em\relax Washington DC: IEEE, 3 2020, pp. 1--6.

\bibitem{Tappeiner2009}
H.~W. Tappeiner, R.~L. Klatzky, B.~Unger, and R.~Hollis, ``{Good Vibrations: Asymmetric vibrations for directional haptic cues},'' in \emph{IEEE World Haptics 2009}, 2009, pp. 285--289.

\bibitem{Brown2021Shape-ChangingVibration}
J.~Brown and I.~Farkhatdinov, ``{Shape-Changing Touch Pad based on Particle Jamming and Vibration},'' in \emph{2021 IEEE World Haptics Conference, WHC 2021}.\hskip 1em plus 0.5em minus 0.4em\relax IEEE Inc., 7 2021, p. 337.

\bibitem{Rahal2020CaringTelemanipulation}
R.~Rahal, G.~Matarese, M.~Gabiccini, A.~Artoni, D.~Prattichizzo, P.~R. Giordano, and C.~Pacchierotti, ``{Caring About the Human Operator: Haptic Shared Control for Enhanced User Comfort in Robotic Telemanipulation},'' \emph{IEEE Trans on Haptics}, vol. 13(1), pp. 197--203, 2020.

\bibitem{Brown2024DesignActuators}
J.~Brown and F.~Bello, ``{Design and Characterisation of Particle Jamming-Based Variable Stiffness Displays using Non-pneumatic Actuators},'' in \emph{IEEE Haptics Symposium (HAPTICS)}, 2024, pp. 379--384.

\bibitem{Brown2026AModulation}
J.~Brown, F.~Bello, and L.~A. Jones, ``{A Multimodal Haptic Interface for Simultaneous Softness and Thermal Modulation},'' in \emph{15th International Conference on Human Haptic Sensing and Touch Enabled Computer Applications, EuroHaptics 2026}.\hskip 1em plus 0.5em minus 0.4em\relax Siena: Springer, 2026.

\end{thebibliography}

\end{document}